\documentclass[10pt,letterpaper]{article}
\usepackage[top=0.85in,left=2.75in,footskip=0.75in,marginparwidth=2in]{geometry}
\usepackage{textcomp}

\usepackage[utf8]{inputenc}

\usepackage{cite}
\usepackage[T1]{fontenc}
\usepackage{indentfirst}
\usepackage{nameref,hyperref}

\usepackage[right]{lineno}

\usepackage{microtype}
\DisableLigatures[f]{encoding = *, family = * }

\raggedright
\usepackage{changepage}
\usepackage{ragged2e}
\usepackage[aboveskip=1pt,labelfont=bf,labelsep=period,singlelinecheck=off]{caption}

\makeatletter
\renewcommand{\@biblabel}[1]{\quad#1.}
\makeatother

\usepackage{lastpage,fancyhdr,graphicx}
\usepackage{epstopdf}
\fancyheadoffset[L]{2.25in}
\fancyfootoffset[L]{2.25in}

\usepackage{color}

\definecolor{Gray}{gray}{.25}

\usepackage{graphicx}

\usepackage{sidecap}

\usepackage{wrapfig}
\usepackage[pscoord]{eso-pic}
\usepackage[fulladjust]{marginnote}
\reversemarginpar

\begin{document}
\vspace*{0.35in}

\begin{flushleft}
{\Large
\textbf\newline{Boost AI Power: Data Augmentation Strategies with unlabelled Data and Conformal Prediction, a Case in Alternative Herbal Medicine Discrimination with Electronic Nose}
}
\newline
\\
Li Liu\textsuperscript{1},
Xianghao Zhan\textsuperscript{1,2},
Rumeng Wu\textsuperscript{1},
Xiaoqing Guan\textsuperscript{1},
Zhan Wang\textsuperscript{1}
Wei Zhang\textsuperscript{1},
Mert Pilanci\textsuperscript{4},
You Wang\textsuperscript{1,*},
Zhiyuan Luo\textsuperscript{3},
Guang Li\textsuperscript{1}
\\
\bigskip
\bf{1} The State Key Laboratory of Industrial Control Technology, Institute of Cyber-Systems and Control, Zhejiang University, Hangzhou 310027, China.
\\
\bf{2} Department of Bioengineering, Stanford University, Stanford, CA, 94305, USA
\\
\bf{3} The Computer Learning Research Center, Royal Holloway, University of London, Egham Hill, Egham, Surrey TW20 0EX, UK
\\
\bf{4} Department of Electrical Engineering, Stanford University, Stanford, CA, 94305, USA
\\
\bigskip
* king\_wy@zju.edu.cn

\end{flushleft}

\section*{Abstract}
\justifying
Electronic nose has been proven effective in alternative herbal medicine classification, but due to the nature of supervised learning, previous research heavily relies on the labelled training data, which are time-costly and labor-intensive to collect. To alleviate the critical dependency on the training data in real-world applications, this study aims to improve classification accuracy via data augmentation strategies. The effectiveness of five data augmentation strategies under different training data inadequacy are investigated in two scenarios: the noise-free scenario where different availabilities of unlabelled data were considered, and the noisy scenario where different levels of Gaussian noises and translational shift were added to represent sensor drifts. The five augmentation strategies, namely noise-adding data augmentation, semi-supervised learning, classifier-based online learning, Inductive Conformal Prediction (ICP) online learning and our novel ensemble ICP online learning (EICP) proposed in this study, are compared against supervised learning baselines, with Linear Discriminant Analysis (LDA) and Support Vector Machine (SVM) as the classifiers. This study provides a systematic analysis of different augmentation strategies. Our novel strategy, EICP, outperforms the others by showing non-decreasing classification accuracy on all tasks and a significant improvement on most simulated tasks (25 out of 36 tasks, $p \leq 0.05$), which demonstrated both effectiveness and robustness in boosting the classification model generalizability. This strategy can be employed in other machine learning applications.

\section*{Keywords}
conformal prediction, data augmentation, electronic nose, herbal medicine, semi-supervised learning

\section*{Introduction}
\label{sec:introduction}
Alternative herbal medicines are valuable in medical research and therapies\cite{1TCM_history}, and their distinct therapeutic effects are associated with specific categories, which can be discriminated with the different emanated volatile organic compounds (VOCs) \cite{38TCM_odors_differ}. However, the subtle difference in the herbal medicine's appearance demands manual classification which relies heavily on the domain expert knowledge and experience \cite{30ZhanW2019ACP_shihu}. To provide a better and safe treatment with the proper alternative herbal medicine, it is worthwhile to develop cheap, effective and reproducible methods for herbal medicine classification\cite{17zhan2018sensors}.

The electronic nose (e-nose), also referred to as artificial olfactory system, mimics the mammalian olfaction for odor pattern recognition\cite{19Diao2020SNA,39E-nose_introduction}. As a cheap and effective analytic system, the e-nose has been applied in environment monitoring\cite{4E-nose_environment,7E-nose_environment}, food production \cite{9E-nose_food,10E-nose_food} and assisted decision making in medical diagnosis\cite{15E-nose_medical,16E-nose_medical}.
	
Although previous research demonstrated e-nose's effectiveness in herbal medicine classification, most proposed algorithms are typically based only on supervised learning, which requires the collection of adequate and high-quality training samples, manual labeling and label verification before training classifiers\cite{17zhan2018sensors,19Diao2020SNA}. However,  collecting adequate training data can be time-consuming, laborious, and costly\cite{17zhan2018sensors}; additionally, label verification requires advanced instruments at civic or provincial pharmaceutical administration agencies. Furthermore, even if the labels of adequate training samples have been verified, sensor drift due to move of the e-nose system from a laboratory environment to a local pharmacy can negatively impact the effectiveness of the training data \cite{32Sensor_drift_ICA,33Sensor_drift_boltzmann_machine}. Therefore, it is hard to get adequate high-quality training data for supervised learning in real-world applications.
	
To alleviate the critical dependency of the supervised learning on labelled data, Li et al.\cite{29_26_zhan_IET_CP+online,30ZhanW2019ACP_shihu,31_Ming_ICP_Brain} applied Conformal Prediction (CP) to improve classification accuracy with augmented data. However, there are some limitations. Firstly, the proposed online-learning method failed to fully consider the real-world situations by using information which usually are not available; for example, true labels were used by Wang et al.\cite{30ZhanW2019ACP_shihu} to validate the effectiveness of conformal prediction. Such true labels would not have been accessible in the online-learning process. Secondly, the data used in previous research\cite{29_26_zhan_IET_CP+online,30ZhanW2019ACP_shihu} were assumed to be homogeneous without considering environment changes or sensor drift as the researchers used the same batch of data collected to validate the online learning protocols. Finally, there was a lack of fair comparison between CP and other possible data augmentation strategies. Researchers did not objectively analyze the performance of the proposed methods over other data augmentation approaches on the same dataset\cite{29_26_zhan_IET_CP+online,30ZhanW2019ACP_shihu,31_Ming_ICP_Brain}.
	
This study investigates the adaptability of e-nose system with data augmentation strategies based on unlabelled data and conformal prediction in classifying 12 categories of alternative herbal medicines (dataset 1), when the inadequacy of training data and their quality mismatch with the on-site data are taken into consideration. Two scenarios are considered to represent real-world applications of e-nose system: 1) there are different ratios of unlabelled on-site data and standard labelled training data, with negligible quality mismatch between them; and 2) there is major data quality mismatch because of sensor drift, but the ratio of unlabelled data and labelled training data is fixed. Then five data augmentation strategies are experimented on the same dataset partition to analyze their efficiency in improving the model generalizability. To validate the experimental results, the same scheme is then applied on another 480 additional validation samples from three categories of alternative herbal medicine collected in  different seasons (dataset 2). Experimental results demonstrate that data augmentation strategies can help the e-nose system adapt to specific applications.

\section*{Methods}
\label{sec:methods}

\subsection*{E-nose system and data collection}

Pattern recognition with e-nose\cite{24_E-nose_mechanism,29_26_zhan_IET_CP+online} can be summarized as the following 2 main steps: 1) the e-nose reacts to VOCs of a sample with the sensor arrays with changing resistance, and the corresponding signals of changing voltage are recorded; and  2) features are extracted from the signals for classification using machine learning algorithms.
    
The main task in this work is to classify 12 categories of alternative herbal medicines (Astragalus, Liquorice, Chinese Angelica, Saposhnikovia Divaricata, Radix Angelicae Pubescentis, Radix Angelicae Dahuricae, Notopterygium Incisum, Codonopsis Pilosula, Radix Bupleuri, Ligusticum Chuanxiong Hort, Radix Peucedani, and Pueraria Lobata), labelled with index from 0 to 11 for simplicity. Fifty samples for each category were collected with an e-nose system consisting of 16 semi-conductive sensors at the State Key Laboratory of Industrial Control Technology, Zhejiang University in the summer of 2017\cite{23_E-nose_mechanism}, and 128 features (for each sensor channel: maximum value, integral value, three sets of both maximum and minimum values of exponential moving average of the signal derivative using different smoothing factors) were extracted according to previous publications\cite{17zhan2018sensors,29_26_zhan_IET_CP+online}.
    
To further validate the effectiveness of different data augmentation strategies, an additional validation experiment is carried out on another dataset (Dataset 2), which consisted of three categories of herbal medicines (Dahurian Angelica, Angelica Biserrata and Pueraria Montana var Lobata, denoted as a, b, c, respectively), with 160 samples in each category. The samples were collected with the same e-nose system in the winter of 2017\cite{18_Feature_engineering}. The same 128 features were extracted for each sample\cite{18_Feature_engineering}.

The dataset is randomly divided into the training, validation, test and active sets, respectively representing the standard training data (given by the research lab), the on-site data with validated labels (may come from validated labels, used to tune model hyperparameters to fit the local environment), the on-site data to be tested (targets of the classification), and the on-site unlabelled data (local data without any validated labels). 
    
To compare different data augmentation strategies under different conditions, two types of scenarios are considered: one for the data quantity mismatch and the other for the data quality mismatch. In the first scenario, the data quantity mismatch is created by setting different levels of unlabelled data availability. It is assumed that the on-site unlabelled data and the standard training data are homogeneous, which are directly partitioned from the total 600 samples. The size of the training set is set to be 60, 120 and 240, and the number of active samples (i.e. the on-site unlabelled data) varies accordingly (300, 240 and 120), as shown in Fig. S1. Therefore, the ratios of the training set and the active set are 0.2, 0.5 and 2, while the validation set and the test set have 120 samples each. This scenario is used to simulate the situation where noises caused by sensor drift are compensated or negligible\cite{43_ICA,44_boltzmann_machine}.

In the second scenario, the data quality mismatch between the lab data (i.e. standard training data) and the on-site data is created by introducing different types and levels of noises. However, the unlabelled data availability is fixed. Once the ratio of the training set and the active set is fixed, taking 0.5 as an example, varying degrees of Gaussian noises or translational shift are added into the on-site data (the validation, the test and the active sets). The Gaussian noises are generated based on the standard derivation ($SD$) of the original datasets according to the following formula: 
    
    \begin{equation}
    \label{noise}
    Noise =c \times S D_{i} \times \delta, \quad \delta \sim N(0,1)
    \end{equation}
 where $i$ denotes the $i$-th feature and $c$ is the coefficient of noise level taking one of the values in ($0.01, 0.03, 0.05$).   
    
For the translational shift noises, the normally distributed $\delta$ in formula (\ref{noise}) is removed and a positive translational shift is added to the features of the on-site data ($c = 0.01, 0.03, 0.05$). Note that negative translational shift are also tested and the similar results are observed, which is not reported due to the page limit. 

The additional validation experiment on dataset 2 follows the same pipeline. In scenario 1, the training set and the unlabelled set are obtained in the ratio of 0.25 (60/240), 0.66 (120/180), and 4 (240/60), with 90 samples in the validation and the test sets respectively. In scenario 2, the Gaussian noises and translational shift are added in the same way as dataset 1 with $c= 0.01, 0.03, 0.05$. Detailed dataset partitions for dataset 2 are shown in Fig. S2.
    
\begin{figure*}
        \centering
		\includegraphics[width=\linewidth]{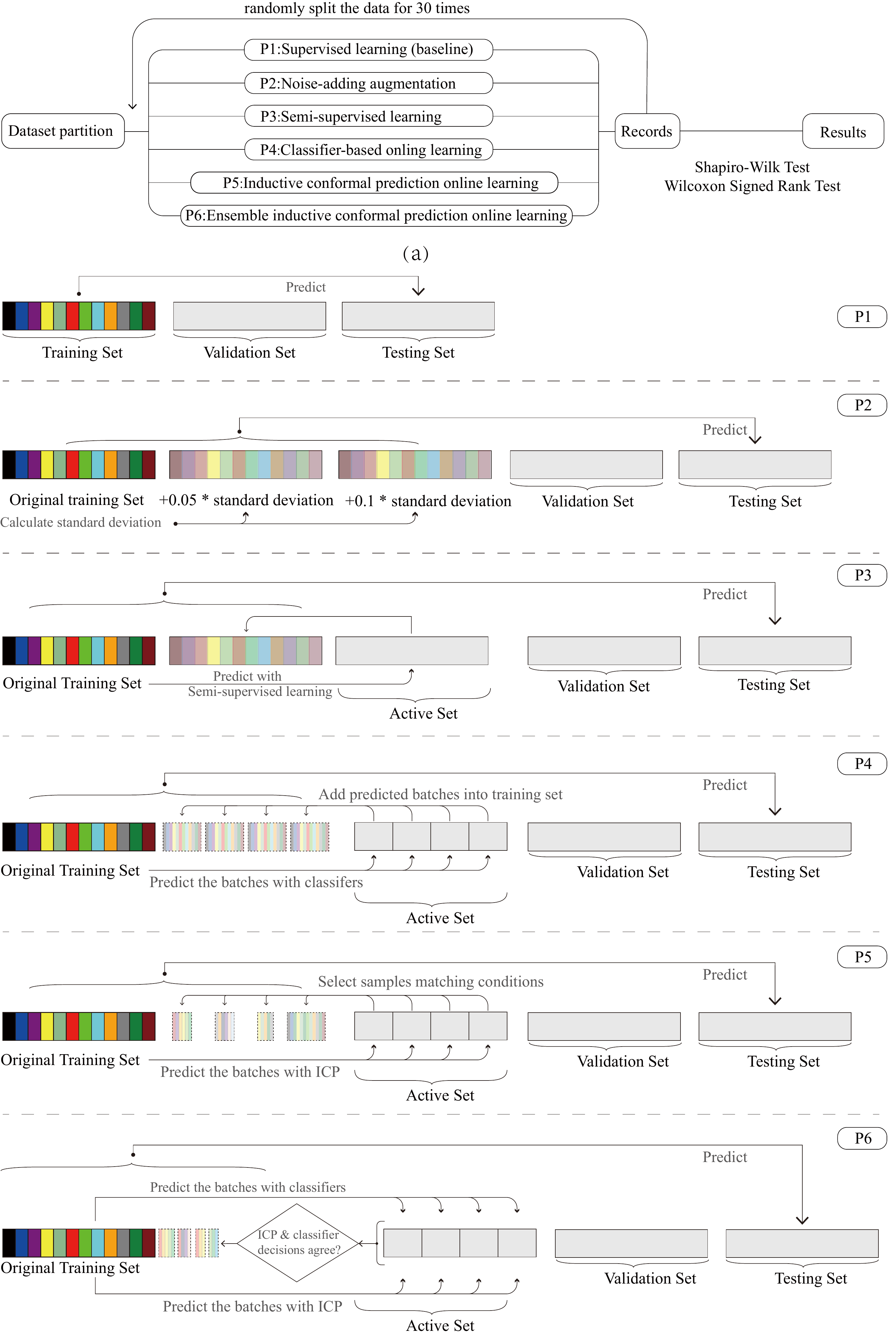}
		\caption{\textbf{The illustration of the six processes in the classification tasks.} (a) The illustration of the six processes in 30 random dataset partitions and statistical significance tests. (b) The illustration of the six processes: supervised learning, noise-adding augmentation, semi-supervised learning, classifier-based online learning, inductive conformal prediction (ICP) online learning, ensemble ICP (EICP) online learning.}
		\label{Fig:process}
	\end{figure*}

After the dataset partition, six processes (P1-P6) which include the baseline supervised learning and five data augmentation strategies are evaluated on the same dataset partition, and their classification accuracy on the test set is then analyzed. For each process in the simulated scenarios, the entire pipeline is repeated 30 times with different random dataset partitions. Wilcoxon signed-rank tests rather than the paired student t-tests are performed to test the statistical significance among the reported classification accuracy results, since the Shapiro-Wilk test rejects the normal distribution of some accuracy results. The whole scheme is illustrated in Fig. \ref{Fig:process}.
 
\subsection{Data augmentation strategies}
In this section, different data augmentation strategies will be presented and discussed.
	\subsubsection{Process 1: Supervised learning}
	This process is set as the baseline process to benchmark the potential accuracy improvement of other data augmentation strategies. In this process,  Linear Discriminant Analysis (LDA)\cite{34.5_LDA} and Support Vector Machine (SVM)\cite{SVM} classifiers are trained on the training set and evaluated on the test set. These two classifiers are chosen because they achieve the highest leave-one-out cross-validation accuracy on the same dataset in the previous publication\cite{17zhan2018sensors}. The validation set is used to tune the hyperparameters, including the kernel function and soft margin penalty coefficient ($C$) in SVM. After that, the hyperparameters are fixed for all the following processes so that the performance difference is only caused by different data augmentation strategies. Our implementation is based on scikit-learn package (version 0.24.0) in Python 3.7\cite{50_sklearn}.

	\subsubsection{Process 2: Noise-adding data augmentation}
	In this process, the training set is augmented with Gaussian noises to improve model generalizability\cite{18_Feature_engineering,P2_effectiveness,P2_effectiveness2}. The augmented data are of the same size as the active set to ensure fair comparison in the sense that in each augmentation strategy, the same amount of data for modeling classifiers is provided. As discussed early, the noises are generated from the standard derivation of the original training samples, with a normal distribution and different scales as defined in formula (\ref{noise}). The noise-added samples are then used to augment the training set. The scales of the Gaussian noises ($c$) are chosen to be 0.05 and 0.1 as recommended in the previous publication using the same feature extraction techniques \cite{18_Feature_engineering}. The test set is predicted based on the augmented training set.

	\subsubsection{Process 3: Semi-supervised learning}
	In this process, three semi-supervised algorithms are applied: label propagation\cite{46_label_propagation}, label spreading\cite{47_label_spreading}, and semi-supervised K-means clustering\cite{48_semi_knn}, implemented using scikit-learn package. Label propagation is a graphical semi-supervised learning method, which assumes that the adjacent sample nodes in the feature space have the same label, and propagates labels through network\cite{53_label_propagation}. Label spreading is the regularized form of label propagation algorithm with affinity matrix\cite{55_label_spreading}. Semi-supervised K-means clustering is developed from K-means algorithm, while it uses the labelled data to determine the cluster numbers and initialize centers. These algorithms are used to assign the labels of active samples and augment the training set. The same classifiers used in Process 1 are modeled with the augmented training set to classify the test samples after the hyperparameters (kernel function and the maximum number of iterations for label spreading and label propagation, the number of neighbors for $k$ nearest neighbors (kNN) kernel, clamping factor ($\alpha$), and bandwidth parameter ($\gamma$) for radial basis function (RBF) kernel\cite{50_sklearn}) are tuned on the validation set. The best semi-supervised learning models are then chosen as the candidates for Process 3. 
	
	\subsubsection{Process 4: Classifier-based online learning }
	This process utilizes the active set via batch-wise online learning: the active set is divided into four equally sized batches, and then each batch is predicted with labels based on the previous training set, then added to augment the training set. Finally, all the unlabelled data are included in the training set with their predicted labels, and the augmented training set is used to classify the test samples.
	
	\subsubsection{Process 5: Inductive conformal prediction online learning}
	In Inductive Conformal Prediction (ICP) online learning, the unlabelled data are predicted by ICP proposed by Vovk et al.\cite{37_ICP,40_CP,41_CP,42_CP}. ICP is a machine learning framework based on the identical and independent distribution assumption and ensures the error rate of the predictions is not higher than a pre-chosen significance level $\epsilon \in [0,1]$. ICP measures the reliability (credibility and confidence) of each prediction result with a quantitative confidence measurement\cite{37_ICP,40_CP,41_CP,42_CP}. The central idea behind ICP is to estimate how well a new test example fits into a distribution of the training set. The nonconformity measurement scores how different a new sample is from the previous observations. The underlying nonconformity measurement can be based on many commonly used classification algorithms. In this study, $k$ nearest neighbors (kNN) is used as the underlying classifier according to the originally proposed ICP method\cite{18_Feature_engineering}.

\begin{figure}
		\includegraphics[width=\linewidth]{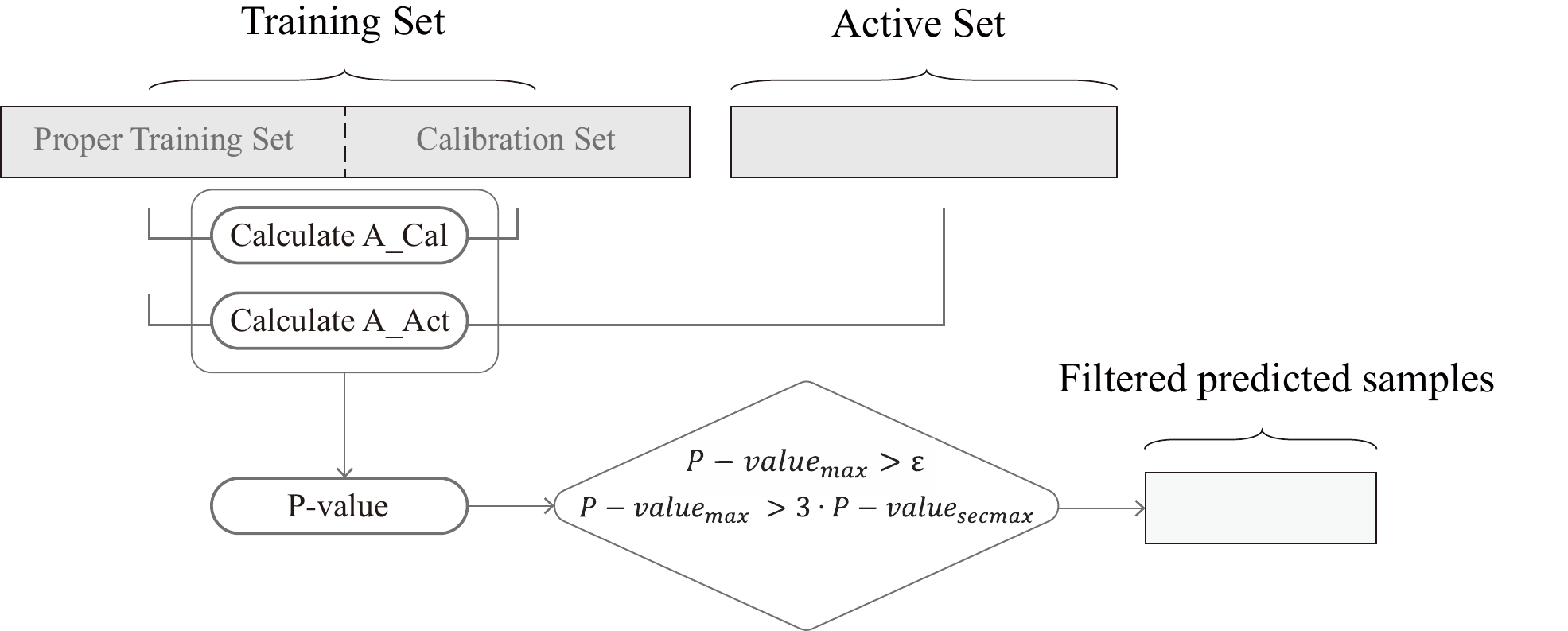}
		\caption{\textbf{The pipeline of ICP online learning.} The nonconformity measurement A\_Cal and A\_Act is calculated based on proper training set. The P-value is calculated and the predictions are filtered by two criteria on P-values to augment the training set.}
		\label{Fig:icp}
	\end{figure}

	The ICP is illustrated in Fig. \ref{Fig:icp}, where the training set with $b$ samples is divided into a proper training set $\left\{X_{1}, X_{2}, X_{3}, \ldots, X_{a}\right\}$ and a calibration set $\left\{X_{a+1}, X_{a+2}, \ldots, X_{b}\right\}(a<b)$. Due to the small size of data in this study, those two sets are combined into one and the entire training set will be used. Based on the training set, the nonconformity measurement of calibration set and active set ($A\_Cal$ and $A\_Act$) are calculated according to the method shown in Fig. \ref{Fig:icp}. The P-values are 
    calculated for all the possible labels of each new sample, indicating the reliability of the prediction. Two criteria based on P-values are designed to ensure that only the samples with reliable predictions will be selected for data argumentation as  shown in Fig.  \ref{Fig:icp} \cite{31_Ming_ICP_Brain}. These two criteria will be discussed later.

	The nonconformity measurement of $i$-th unlabelled sample with the hypothesised label $y$, $\alpha_{i}^{y}$, is calculated by the following formula, where $d\left({x}_{{i}}, {x}_{{m}}^{{y}}\right)$ denotes the distance between the new sample and $k$ nearest samples with the same label $y$ and $d\left({x}_{{i}},{x}_{{m}}^{{!y}}\right)$ denotes the distance between the new sample and k nearest samples with a different label (which is denoted by $!y$ ):
	
	\begin{equation}
         \alpha_{i}^{y}=\frac{\sum_{m=1}^{k} d\left(x_{i}, x_{m}^{y}\right)}{\sum_{l=1}^{k} d\left(x_{i}, x_{l}^{! y}\right)}
    \end{equation}

	The P-values for a new sample $x_i$ are computed as follows. Here, $p_{i}^{y}$ denotes the P-value of $x_i$ for a possible label $y$ in the label space, $\alpha_{j}^{y}$ denotes the nonconformity measurement of $j$-th sample with the label $y$ in the calibration set, and $\alpha_{i}^{y}$ denotes the nonconformity measurement for a possible label $y$ of $i$-th sample in the active set.
	\begin{equation}
        p_{i}^{y}=\frac{| \{j=a+1, \ldots, b \}:  \alpha_{j}^{y} \geq \alpha_{i}^{y} \mid+1}{b-a+1}
    \end{equation}
    
    To ensure the quality of the augmented data, the output predictions need to satisfy several reliability requirements before adding to the training set. In the ICP method, the credibility of a prediction is the largest P-value and the confidence is 1 minus the second-largest P-value\cite{29_26_zhan_IET_CP+online,30ZhanW2019ACP_shihu,37_ICP}. Accordingly,  two thresholds are set in this study as shown in Fig. \ref{Fig:icp}. The first threshold $\varepsilon$ set on the maximum P-values enables the output prediction to achieve at least a minimum credibility\cite{28_CP,31_Ming_ICP_Brain,37_ICP}\cite{49_ICP}. The second threshold requires the largest P-value to be at least three times higher than the second largest P-value for a sample. Setting those two thresholds means both the credibility and the confidence measures of predictions are assessed, and only the samples satisfying both two requirements are used for augmentation, taking the predictions with the highest P-value as their corresponding labels. In this process, the hyperparameters $k$ (the number of nearest neighbors) and $\varepsilon$ (the significance level of conformal prediction) are tuned based on the classification accuracy on the validation set. The process is run in a batch-wise manner as in Process 4.

	\subsubsection{Process 6: Ensemble inductive conformal prediction online learning (EICP)}
	
	To incorporate the knowledge of both the classifier and the conformal predictor, we propose ensemble ICP online learning (EICP/Process 6). In this process, the intersection of the prediction results given by Processes 4 and 5 are taken, where the probability and reliability of predictions are respectively modeled by classifiers and ICP. The combination of predictions is conducted as follows: for a new batch in active set with $n$ samples, the outputs of Process 4 and Process 5 are denoted as$\left\{\left(X^{4}, Y^{4}\right)\right\}$ with the size of $n$ and $\left\{\left(X^{5}, Y^{5}\right)\right\}$ with the size of $m$ $(n \ge m)$, as Process 4 adds all active samples in the augmentation but Process 5 filters the active samples based on the two criteria. For each sample in the new batch, if its predictions given by Processes 4 and 5 are matched, denoted as $\left\{\left(x_{i}, y_{i}\right) \mid x_{i} \in X^{4} \cap X^{5}, y_{i}^{4}=y_{i}^{5}, i=1, \ldots, m\right\}$, it will be added to the training set augmentation in Process 6.

\section{Result}
    \subsection{Data visualization}
\begin{figure}
		\includegraphics[width=\linewidth]{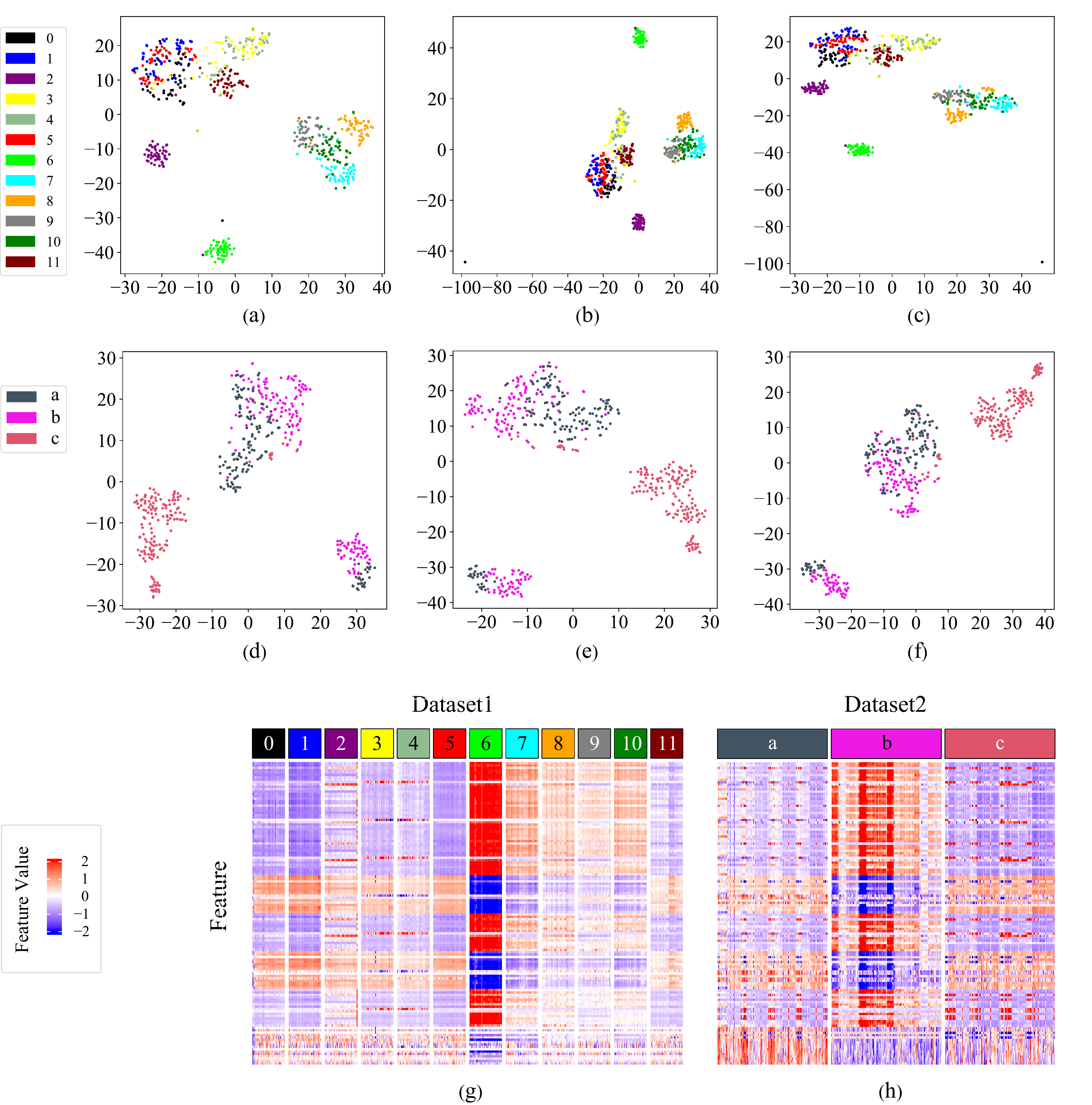}
		\caption{\textbf{The visualization of the dataset 1 and dataset 2 with t-SNE and heatmap.} (a) The t-SNE visualizations of the dataset 1 without noises, (b) with Gaussian noises ($c=0.05$), (c) with translational shift ($c=0.05$). (d) The t-SNE visualizations of the dataset 2 without noises, (e) with Gaussian noises ($c=0.05$), (f) with translational shift ($c=0.05$). The heatmap visualization of the dataset 1 (g) and dataset 2 (h).}
		\label{Fig:visualization}
	\end{figure}

    The distribution of dataset 1  is visualized in Fig.  \ref{Fig:visualization}(a)-(c) using t-distributed Stochastic Neighbor Embedding (t-SNE)\cite{t-SNE}, and Fig. \ref{Fig:visualization}(g) with feature heatmap, while the t-SNE visualization and heatmap for dataset 2 are shown in Fig. \ref{Fig:visualization}(d)-(f) and (h). It can be seen that the samples of different categories scattered into different clusters in t-SNE plots, which indicated the classification feasibility. Fig. \ref{Fig:visualization}(b) and (c) show that the samples added with Gaussian noise ($c=0.05$) and translational shift ($c=0.05$) tend to be mixed in t-SNE plots. The heatmap\cite{heatmap} in Fig. \ref{Fig:visualization}(g) depicts that different categories have their specific feature patterns but many categories have similar features, such as categories 0, 1, 3 and 5.

	\subsection{Performance of data augmentation strategies}
	
	 In this study, 36 classification tasks are designed to evaluate the effectiveness of different augmentation strategies. These tasks are set as the combinations of 2 datasets (dataset 1 and dataset 2), 2 classifiers (LDA and SVM), and 9 simulated training inadequacy scenes: 3 different ratios of dataset partition in scenario 1, 3 different levels of Gaussian noises and 3 different levels of translational shift in scenario 2. The detailed experimental results are presented in Fig. S3-S6, displaying the  performance of different strategies on the test set in 36 tasks.

\begin{figure*}
	    \centering
		\includegraphics[scale=0.6]{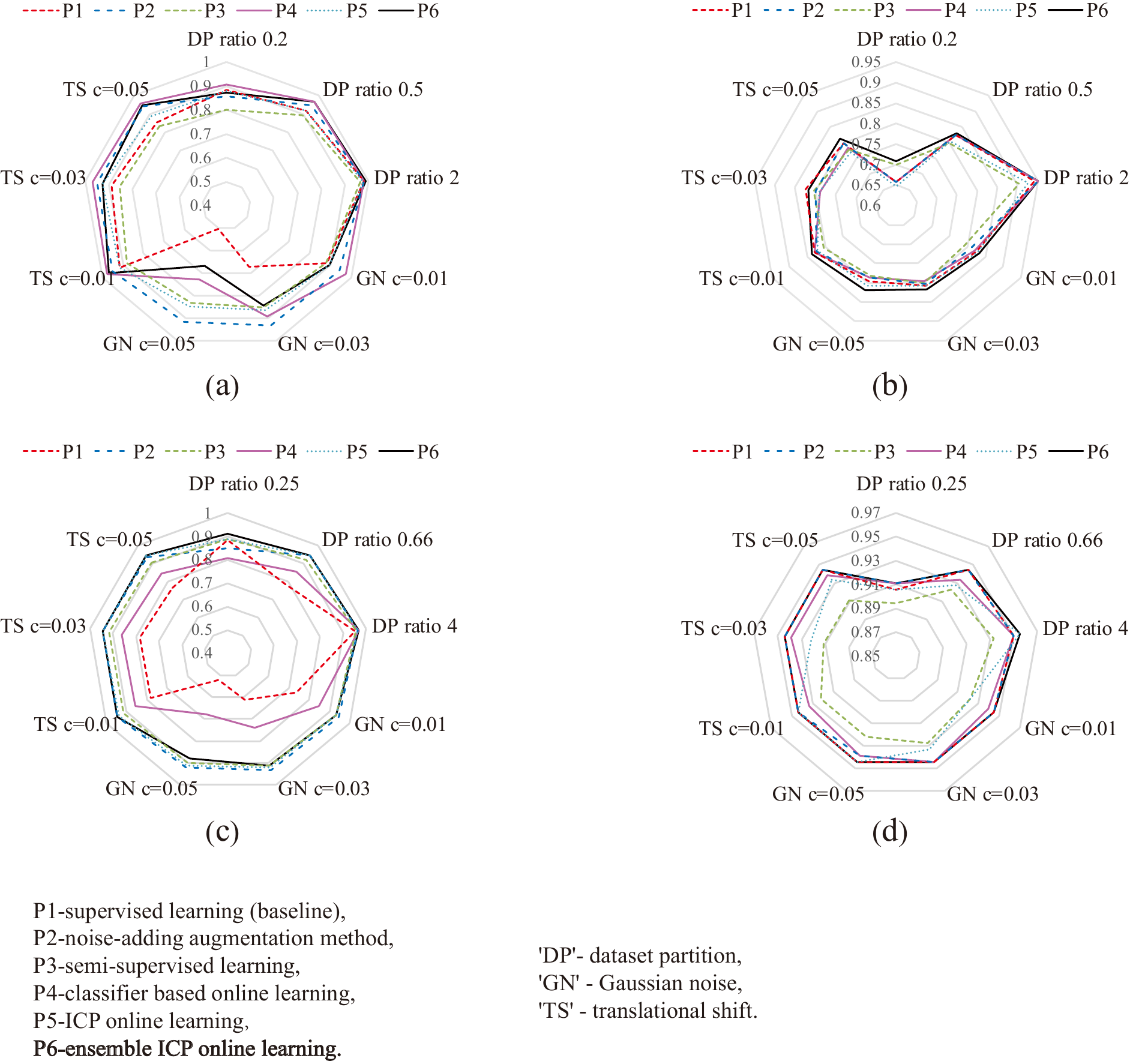}
		\caption{\textbf{The summary of effect on classification accuracy of six processes in nine tasks on two datasets.} The median classification accuracy of six processes on dataset 1 with LDA (a), dataset 1 with SVM (b), dataset 2 with LDA (c), dataset 2 with SVM (d).}
		\label{Fig:radar_plot}
	\end{figure*}

	 To visually summarize the performance of different data augmentation strategies in 36 tasks, the median classification accuracy of the 30 repeats of random dataset partitions is represented in the radar charts in Fig. \ref{Fig:radar_plot}(a)-(d). From these charts, there are at least one strategy that improved the classification accuracy with LDA and achieved non-decreasing classification accuracy with SVM when compared with the baseline process (Process 1) in each of the classification tasks. Furthermore,  our novel Process 6 performs consistently well in the majority of the tasks.
	
   The quantitative summary of the performance of different augmentation strategies is given in Table \ref{table_counting},  which shows their effect on classification accuracy, and the counts of statistically significantly increasing (+), non-increasing (=), and decreasing (-) accuracy compared with the baseline in 36 tasks.
   
    \begin{table}[htbp]
\renewcommand\arraystretch{1.5}

\caption{The counts of statistically significant accuracy changes of the five data augmentation strategies in 36 tasks, compared with supervised learning (Process 1)}
\label{table_counting}
\scalebox{0.77}{
\begin{tabular}{lcccc}
\hline
                                    & \multicolumn{1}{l}{+  (p\textless{}0.05)} & \multicolumn{1}{l}{= (p\textgreater{}0.05)} & \multicolumn{1}{l}{- (p\textless{}0.05)} & Total \\ \hline
P2: noise-adding augmentation method & 15                                        & 16                                          & 5                                        & 36    \\
P3: semi-supervised learning         & 10                                        & 7                                           & 19                                       & 36    \\
P4: classifier based online learning & 17                                        & 17                                          & 2                                        & 36    \\
P5: ICP online learning              & 12                                        & 18                                          & 6                                        & 36    \\
\textbf{P6: ensemble ICP online learning (proposed)}     & 25                                        & 11                                          & 0                                        & 36    \\ \hline
\end{tabular}}
\end{table}
   
   Note that the proposed novel data augmentation strategy in this study, EICP (Process 6), is the only strategy that showed a non-decreasing accuracy in all the tasks. It achieves statistically significant accuracy improvement on 25 tasks (p\textless{}0.05) and non-decreasing classification accuracy on the remaining 11 tasks (p\textgreater{}0.05). The detailed experimental results on different scenarios and datasets are discussed below.

\begin{figure*}
	    \centering
		\includegraphics[width=\linewidth]{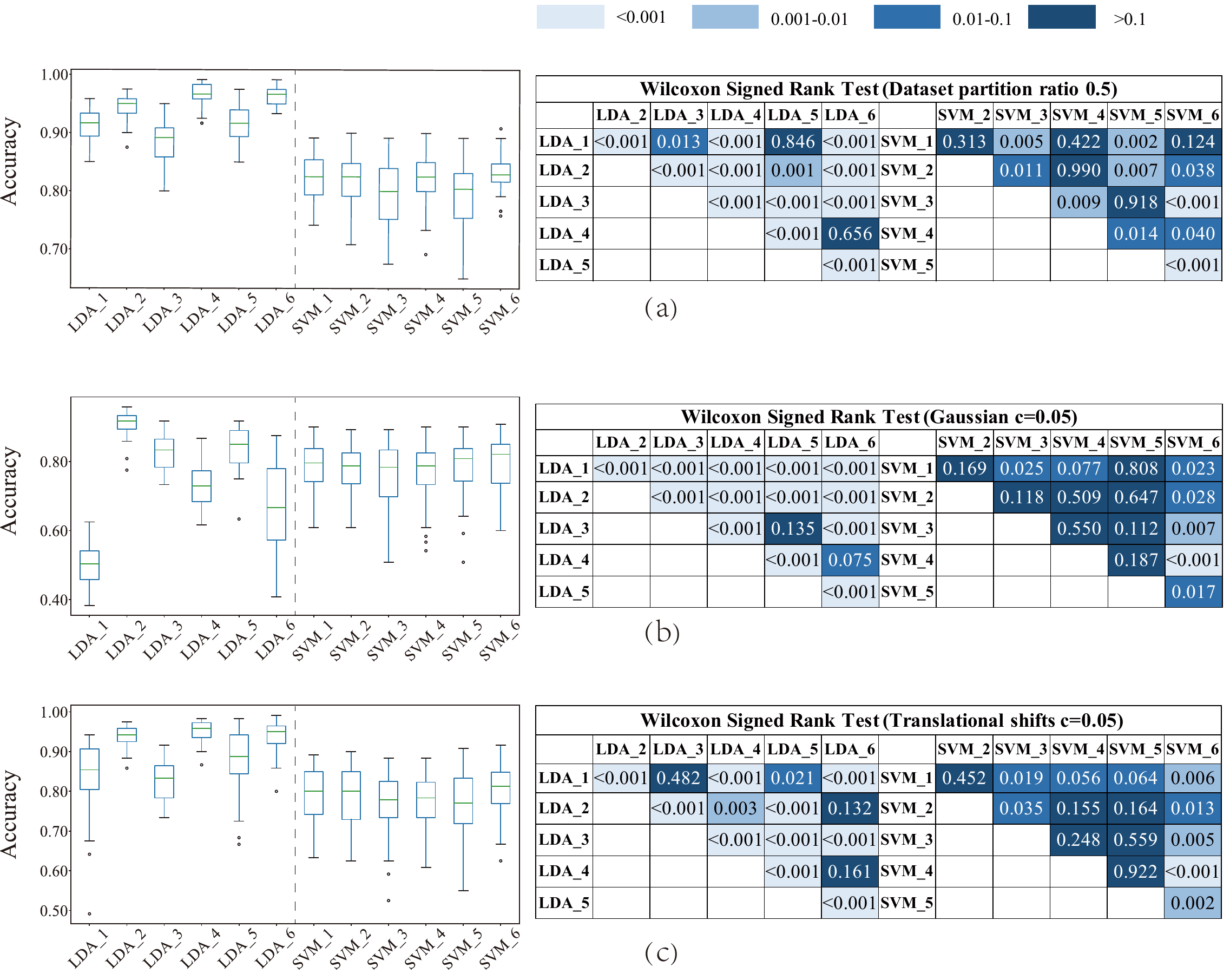}
		\caption{\textbf{The classification accuracy and the statistical significance given by Wilcoxon signed-rank tests of six processes in scenario 1 and 2 (dataset 1).} The figures denote the dataset partition with a ratio of 0.5 in scenario 1 (a), and two types of noisy conditions in scenario 2: Gaussian noises with c=0.05 (b) and translational shift noises with c=0.05 (c).}
		\label{Fig:dataset 1}
	\end{figure*}

	The performance of different strategies on dataset 1 is presented in Fig. \ref{Fig:dataset 1}. The results of three tasks are displayed in Fig. \ref{Fig:dataset 1} as examples, where dataset partition with a ratio of 0.5(120/240) in scenario 1, Gaussian noise (c=0.05) and the translational shift (c=0.05) in scenario 2. All results on dataset 1 can be found in Fig. S3 (for scenario 1) and Fig. S5 (for scenario 2). \\
	\indent In scenario 1, comparing the LDA-based processes with the baseline supervised learning (Process 1) indicates that there is a general increase in the classification accuracy in Processes 4 ($p \leq 0.05$) and 6 ($p \leq 0.05$ on the ratios of 0.5 and 2, $p = 0.508$ on the ratio of 0.2), while Process 3 has inferior accuracy  ($p \leq 0.05$). As for SVM-based processes, Processes 4 and 6 show non-decreasing accuracy. Process 5 has generally inferior accuracy than the baseline supervised learning ($p \leq 0.05$ on the ratios of 0.5 and 2, $p = 0.773$ on the ratio of 0.2).

	For scenario 2, under both Gaussian noises and translational shift, the supervised learning accuracy of the LDA models gradually decreases with the increase of noise level, which indicates that the generalizability of the supervised learning is negatively influenced by the data quality mismatch. The majority of augmentation processes show their effectiveness in improving model generalizability under both types of noises. For example, Processes 4 and 6 show a general increase in the classification accuracy ($p \leq 0.05$). Process 2 performs well in improving classification accuracy under Gaussian noises as it adopts Gaussian noises in data augmentation ($p \leq 0.05$). It also provides improvement under translational shift ($p \leq 0.05$).  Process 3 achieves non-decreasing classification accuracy under Gaussian noises and shows significantly improved classification accuracy in noisier conditions ($p \leq 0.05$, $c=0.03, 0.05$), which is better than that in scenario 1. However, Process 3 suffers significant decrease in accuracy under translational shift ($p \leq 0.05$, $c=0.01, 0.03$).  For SVM models, higher robustness to noises is evident for Process 1 under different levels of noises. Process 6 obtains a statistically significant increase in classification accuracy ($p \leq 0.05$) at three levels of Gaussian noises and three levels of translational shift, while Process 4 demonstrates non-decreasing classification accuracy in all those tasks. Process 5 shows a general decrease in high noise conditions ($p \leq 0.05$, $c=0.03, 0.05$ for translational shift) and Process 3 exhibits a general decrease under all noise conditions ($p \leq 0.05$).

\begin{figure}
		\includegraphics[width=\linewidth ]{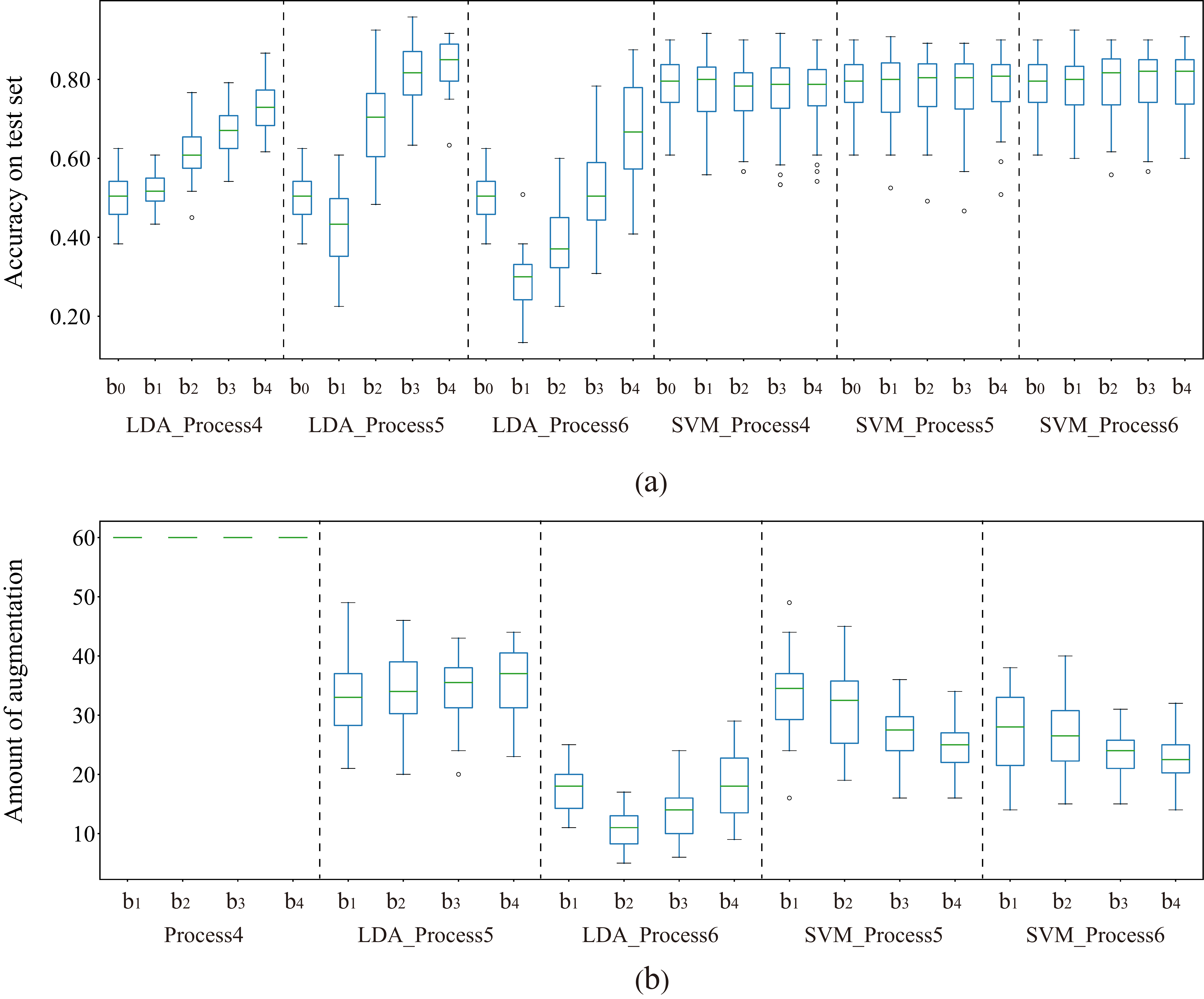}
		\caption{\textbf{The amount of augmented data and the accuracy on the test set in scenario 2 with Gaussian noises ($c=0.05$) (dataset 1 with the ratio of 0.5 (120/240))}:  (a) denotes the  the accuracy in the online learning process. (b) shows the amount of augmented data in the online learning process where b1, b2, b3 and b4 denotes batch 1, batch 2, batch 3, and batch 4, respectively.}
		\label{Fig:augmentation growth}
	\end{figure}

	To analyze the effect of augmented data in Processes 4, 5 and 6, the accuracy of their predictions and the amount of augmented data in each batch are presented in Fig. \ref{Fig:augmentation growth}. The accuracy exhibits a notched tendency, see Fig. \ref{Fig:augmentation growth}(a) as the first batch is added into the original training set, the accuracy drops probably due to the limited classifier knowledge in the initial stage of online learning. As more batches added, the accuracy surpassed the initial level, which indicates that the data augmentation improves the generalizability of the classifiers. The amount of augmented data decreases from Process 4 to Process 6, due to the stricter filtering criteria used in Process 6.

\begin{figure*}
		\centering
		\includegraphics[width=\linewidth]{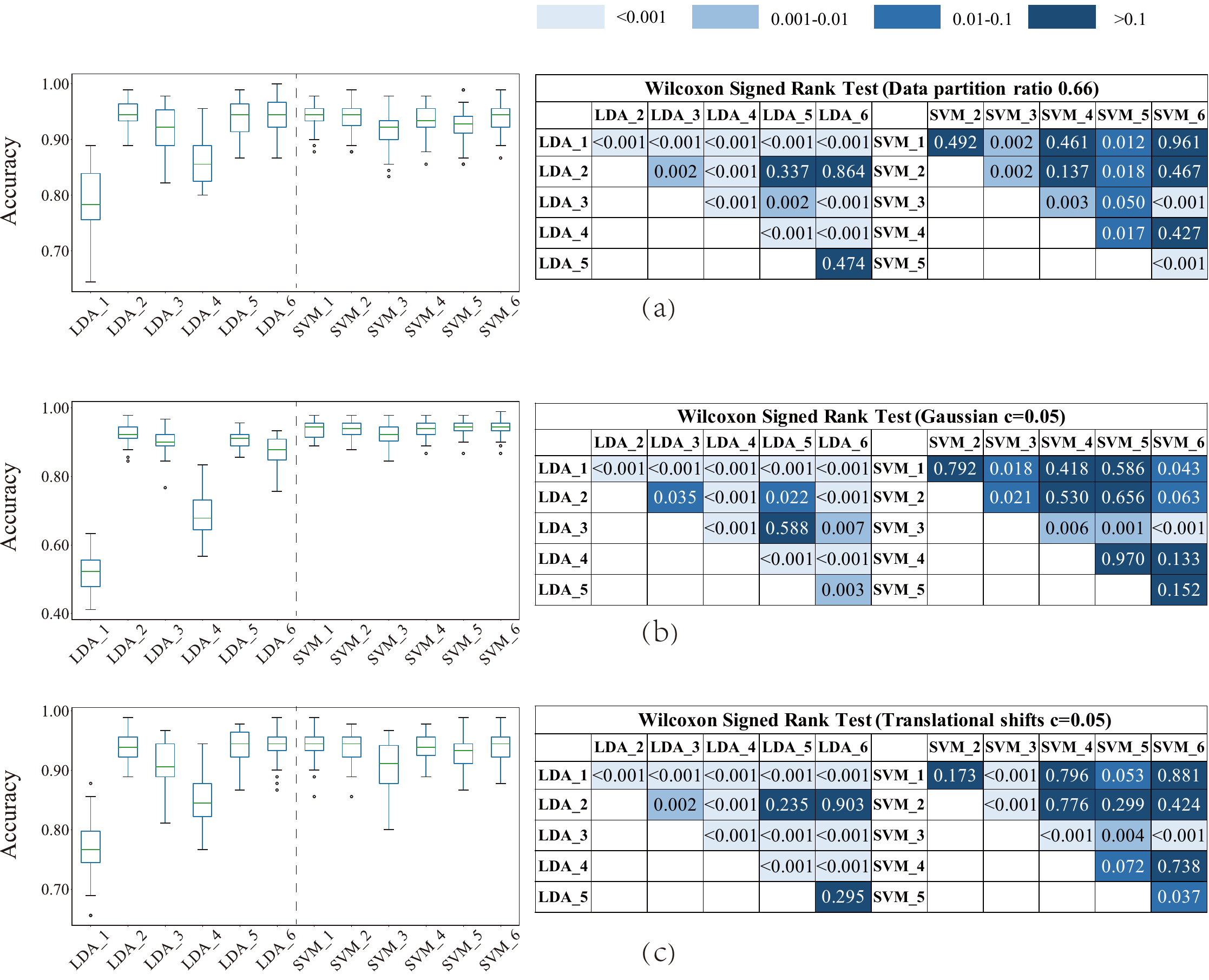}
		\caption{\textbf{The classification accuracy and the statistical significance given by Wilcoxon signed-rank tests of six processes in scenarios 1 and 2 (dataset 2).} The figures denote the dataset partition with a ratio of 0.66 in scenario 1 (a), and two types of noisy conditions in scenario 2: Gaussian noises with c=0.05 (b) and translational shift noises with c=0.05 (c).}
		\label{Fig:dataset 2}
	\end{figure*}

	To validate the effectiveness of the data augmentation strategies, an additional validation experiment on dataset 2 is carried out. In scenario 1, the ratio of the training set and the active set is set to 0.25 (60/240), 0.66 (120/180), and 4 (240/60). In scenario 2, the scales of Gaussian noises and translational shift are set to be 0.01, 0.03 and 0.05. The results of scenario 1 and scenario 2 are shown in Fig. S4  and Fig. S6, respectively. Fig. \ref{Fig:dataset 2} presents the results of three tasks as examples where (a) data partition with a ratio of 0.66 in scenario 1, (b) Gaussian noise with c=0.05 in scenario 2, and (c) the translational shift with c=0.05 in scenario 2.\\
	\indent In scenario 1 with LDA as the base classifier, our novel augmentation strategy (Process 6) improved the classification accuracy ($p \leq 0.05$) on all three different ratios. Process 3 showed non-decreasing accuracy. Process 4 achieved an increase in accuracy on the ratios of 0.66 and 4 ($p \leq 0.05$) but a decrease in accuracy on the ratio of 0.25 ($p \leq 0.05$). Processes 2 and 5 improved on the two ratios ($p \leq 0.05 $ on ratio of 0.66 and 4). In SVM-based processes, the strategies generally have non-decreasing effect. Process 3 generally showed decreases in classification accuracy than the baseline supervised learning.\\
	\indent In scenario 2, all LDA-based processes provided significant improvement in classification accuracy ($p \leq 0.01$) under both Gaussian noises and translational shift, while the baseline supervised learning showed a gradual accuracy decrease with the increasing noise levels. As Fig. \ref{Fig:dataset 2} shows, for SVM-based processes, Process 1 maintained higher robustness under Gaussian noises and translational shift, and Processes 2, 4 and 6 showed non-decreasing effect on classification accuracy, with Process 6 achieving a statistically significant improvement on Gaussian noise ($c=0.05$, $p \leq 0.05$). However, Process 3 gave generally decreasing accuracy in all levels of Gaussian noises and translational shift ($p \leq 0.05$).
	
	\section{Discussion}
   To alleviate the critical dependency on high-quality training data, this study showed that the data augmentation strategies with unlabelled data could improve the classification accuracy with statistical significance (Fig. \ref{Fig:dataset 1},Fig. \ref{Fig:dataset 2}) under various circumstances, where different types and levels of training data quality scenarios were simulated. The effectiveness and robustness of five different data augmentation strategies were experimented and analyzed on two datasets in the tasks of classifying different categories of alternative herbal medicine with electronic nose.

    The first contribution of our study is the proposed new augmentation strategy, namely EICP (Process 6). With a systematic comparison with other data augmentation strategies under different scenarios, it outperformed the other strategies in: 1) showing non-decreasing accuracy in all the simulated tasks (Table \ref{table_counting}); 2) obtaining statistically significant accuracy improvement in most of the tasks; and 3) generally achieving the highest median classification accuracy in the majority of tasks (Fig. \ref{Fig:radar_plot}). The reason why our data augmentation approach outperformed other strategies maybe due to that Process 6 filtered the predictions of the unlabelled data with a stricter dual-criterion, combining the class-probability prediction information from the basic classifiers (LDA and SVM) and the prediction reliability information given by the ICP online learning. As a result, the quality of augmented samples were maintained during the online learning process.
    
    Besides the proposed Process 6, the other data augmentation strategies (Process 2-Process 5) were not performing stably across different tasks. In contrast to Process 6, the conventional ICP online learning (Process 5) only took the prediction reliability information into consideration, and the classifier-based online learning (Process 4) only relied on the class-probability prediction information. This means Processes 4 and 5 are more tolerant of the quality of augmented data and include more unlabelled data for augmentation, which showed good performance in some specific tasks when the quantity of the initial training data was relatively inadequate, but also suffered from statistically significant decrease in accuracy in certain tasks. Noise-adding augmentation (Process 2) generally improved the classification accuracy in many tasks (15 out of 36 tasks), especially under the Gaussian noise conditions. However, the way Process 2 augmented the initial training set was also by adopting Gaussian noises, which would neutralize the effect of Gaussian noises in scenario 2 which intended to simulate data mismatch. In other forms of noises, its effectiveness could not be guaranteed, as is shown by the statistically significant decrease in accuracy in some tasks (5 out of 36 tasks) in Fig. \ref{Fig:radar_plot}. For the semi-supervised learning (Process 3), it led to more significant decrease cases (19 out of 36 tasks) than the improvement cases (10 out of 36 tasks).

    As the data augmentation strategies produce varying performances in different tasks, it is important to help users choose the appropriate data augmentation strategies according to their specific situations. In general, SVM was more stable with different tasks and processes, but LDA showed remarkable accuracy improvements in some cases. For LDA-based models, when the supervised information in the initial training set is relatively inadequate due to either the limited labelled data or data mismatch caused by sensor drift, the quantity of the augmented data plays an increasingly important role. In this case, conventional ICP online learning (Process 5) that includes more active samples can offer more noticeable improvement in the classification accuracy. This can be observed in the tasks which have low ratios of labelled data (in scenario 1) and high noises (in scenario 2) on both datasets 1 and 2. However, when the initial supervised information are relatively adequate, the quality of augmented samples tends to be more important than their quantity. In this case, the novel EICP (Process 6) outperforms the other augmentation strategies in boosting the classification accuracy.  This can be seen in the tasks where there are high ratios of labelled data (in scenario 1) and low noises (in scenario 2) on both dataset 1 and dataset 2. For SVM-based models, it is clear from Fig. \ref{Fig:radar_plot} that the proposed EICP (Process 6) is effective and robust while the other augmentation strategies perform poorly in the same tasks. \\
   \indent The reason why LDA and SVM models acted differently under same tasks and augmentation strategies might be their principles in making predictions. LDA provides the Bayes optimal classifiers when the class conditional distributions are multivariate Gaussian with shared covariance matrices. However, this assumption is often violated with real data, and the LDA models are not necessarily optimal in such cases. On the other hand, SVM only learns a linear decision boundary and therefore it is more robust to outliers and non-Gaussian data distributions. With stricter criteria in filtering augmentation samples, the proposed EICP (Process 6) ensured the augmentation a higher conformity with the original training set, and therefore it was the only strategy that achieved clear improvement in LDA-based tasks and showed non-decreasing classification accuracy in all SVM-based tasks.
    
    Our second contribution is a fair and systematic comparison of different data augmentation strategies. The previous studies have shown that conformal prediction online learning\cite{31_Ming_ICP_Brain,37_ICP} was effective in improving classification accuracy. However, the online learning approaches have only been compared with the baseline supervised learning (Process 1 of this study) in the previous studies while no comparison has been made across different data augmentation strategies, leaving it unknown whether the stated process was superior to other strategies under different circumstances. In addition, the previous studies only investigated the noise-free conditions without considering the data quality mismatch, because the studies simply partitioned the data collected under similar conditions (i.e., temperature, humidity) into training samples and active samples. Furthermore, some of the previous work even directly used the true labels of the active samples to augment the training data\cite{30ZhanW2019ACP_shihu}, which unfairly placed the proposed method in a favorable condition, as the labels of the unlabelled data are typically unknown in the real-world applications. In this study, the multiple data augmentation strategies have been compared with supervised learning (Process 1) as the baseline. To comprehensively evaluate the effect and robustness of different augmentation strategies,  two scenarios (noise-free vs noisy) based on two popular classifiers (LDA and SVM) were considered. Furthermore, the same protocol has also been applied on an additional validation dataset collected in a different season.

    From the viewpoint of algorithms,  these data augmentation strategies and the newly proposed EICP can be utilized in broader ranges of application domains, particularly where there are obstacles in collecting adequate high-quality training data. Examples include medical imaging where the data labeling can be laborious for radiologists\cite{51_medical_radiology}, the application of clinical natural language processing (NLP) where there can be millions of unlabelled notes in the mining of clinical texts\cite{53_NLP_clinical_texts} and the application of optimizing finite element (FE) modeling where each training sample can take hours to collect as the FE models are mostly complicated\cite{52_head_model} but the unlabelled data can be relatively easy to collect or simulate. The data augmentation pipeline proposed in this study can be readily applied in these areas to improve model generalizability with inadequate training data.

	There are some limitations of this study. Firstly, only the Gaussian noises and translational shift were covered in the features and this might be unrealistic in real-world applications to model sensor drift and environment changes from lab to on-site data. This might give noise-adding data augmentation (Process 2) biased advantage as it adopted the same noise-adding method, which may neutralize the effect caused by Gaussian noises intended for data mismatch. Further studies on modeling the actual noises with sensor drift and environment changes need to be investigated. 
	
	Secondly, the dataset in this study was relatively small for ICP. In ICP, there should ideally be a separate proper training set to calculate the nonconformity measurement and a calibration set to calculate the P-values. However, due to the limited amount of data available,  these two sets and directly were combined and all training samples used to fulfil the dual roles. This may lead models to be slightly more likely to suffer from overfitting. The small amount of data may also result in weak confidence measure in the conformal prediction, as the dataset cannot provide enough information to show statistical significance reflected by the P-value calculation in the ICP. Consequently, the active samples which may be potentially beneficial to improving model generalizability were likely to be dropped because the prediction confidence measure could not meet the ICP thresholds specified in Process 5. In the future, more modifications of the ICP and Process 6 can be made to adapt the algorithms to small datasets for better classification improvement. For instance, a bootstrapping method can be included in the nonconformity measurement calculation to ensure more robust results for small datasets.

	\section{Conclusion}
	This study aimed to improve the generalizability of the electronic nose system in discriminating different categories of alternative herbal medicines with five different data augmentation strategies to reduce dependency on high-quality training data in the real-world applications. Two scenarios on two different datasets were considered: 1) the noise-free scenario with the different ratios of training samples and unlabelled data (different availability of unlabelled data) and 2) the noisy scenario with different levels of Gaussian noises and translational shift (data quality mismatch). In both scenarios,  the effectiveness and robustness of five data augmentation strategies were systematically analyzed with LDA and SVM as the classifiers and the classification accuracy as the metric. The experimental results showed: 1) our novel augmentation strategy, ensemble inductive conformal prediction online learning (EICP) outperformed other strategies in most of the times with statistically significant improvement in classification accuracy in the tasks, and it was also the only strategy that showed a non-decreasing classification accuracy in all the tasks and with all classifiers; 2) there was at least one data augmentation strategy on each task that significantly improved the classification accuracy with LDA and led to non-decreasing accuracy with SVM; and 3) the augmentation strategies performed differently in the tasks and recommendations on how to choose the appropriate augmentation strategies adaptively according to different scenarios were discussed. This study provides users with proper augmentation strategies, particularly the novel EICP strategy, to enable the e-nose classification system to be more adaptive to the local environment via unlabelled data utilization and could be applied further to many other machine learning applications.  
	

\newpage

\section*{Supporting Information}
Supplementary matrials are as follows:

\setcounter{figure}{0}
\renewcommand{\thefigure}{S\arabic{figure}}


\begin{figure}[htbp]
\centering
\includegraphics[scale=0.5]{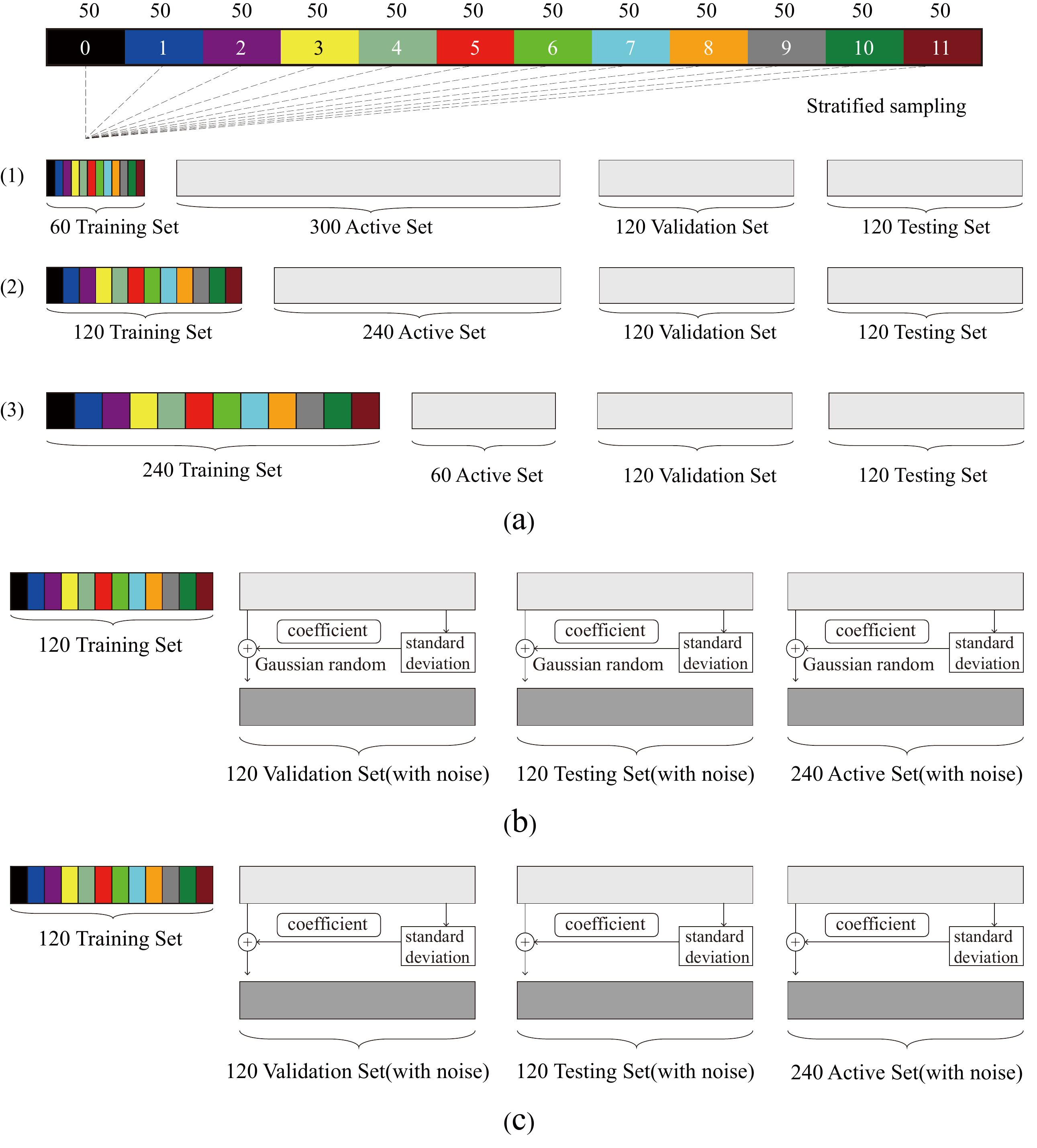}
\caption{\textbf{The illustration of the dataset partition in two simulated scenarios.(dataset 1)} (a) The dataset partition in the first scenario with three different ratios of training samples and active samples. (b) The dataset partition in the second scenario with three different levels of Gaussian noise2. (c) The dataset partition in the second scenario with three different levels of translational shift.}
\label{Fig:dataset1_partition}
\end{figure}

\begin{figure}[htbp]
\centering
\includegraphics[width=\linewidth]{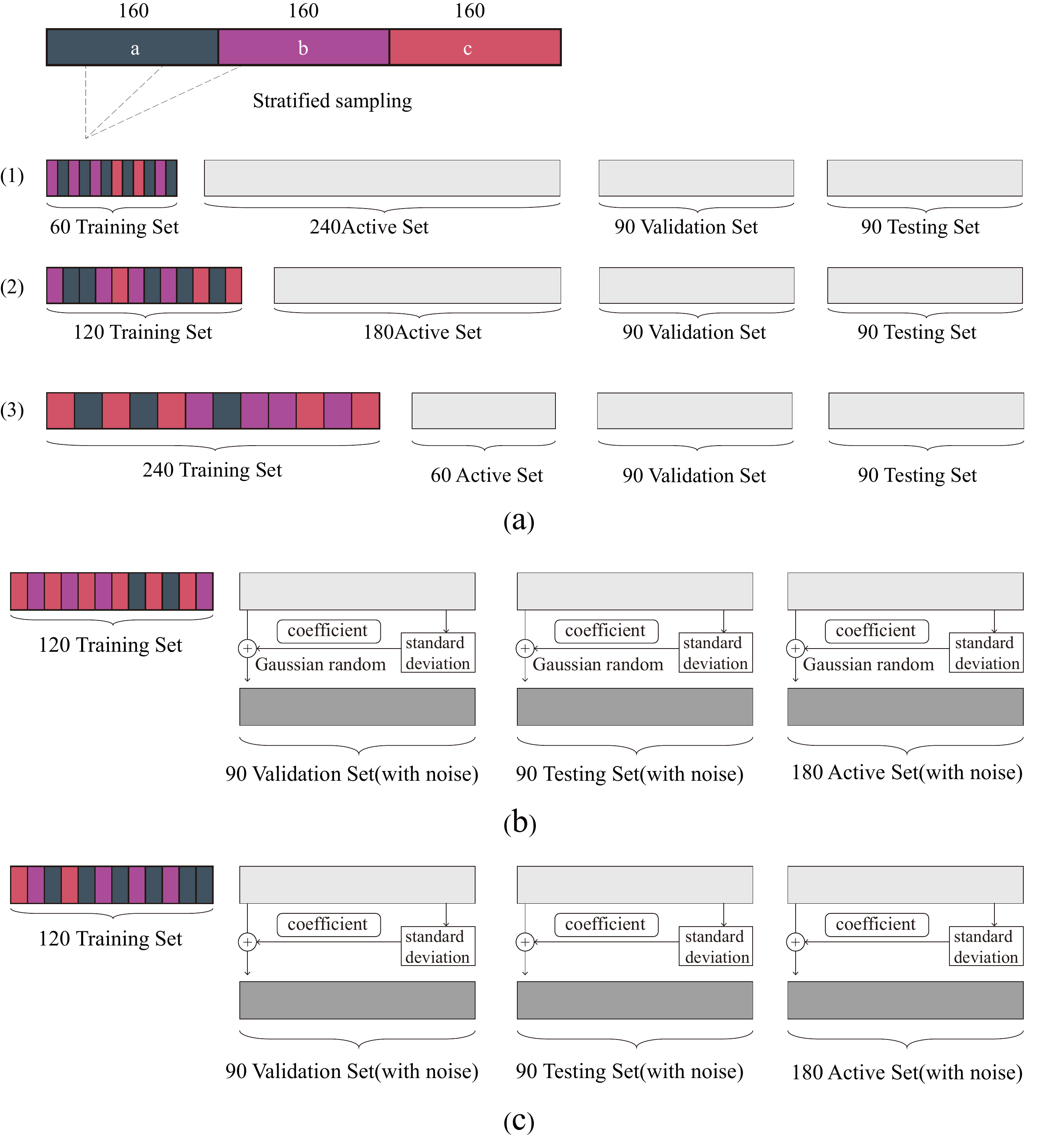}
\caption{\textbf{The illustration of the dataset partition in two simulated scenarios(dataset 2).} (a) The dataset partition in the first scenario with three different ratios of training samples and active samples. (b) The dataset partition in the second scenario with three different levels of Gaussian noise2. (c) The dataset partition in the second scenario with three different levels of translational shift.}
\label{Fig:dataset2_partition}
\end{figure}

\begin{figure}[htbp]
\centering
\includegraphics[width=\linewidth]{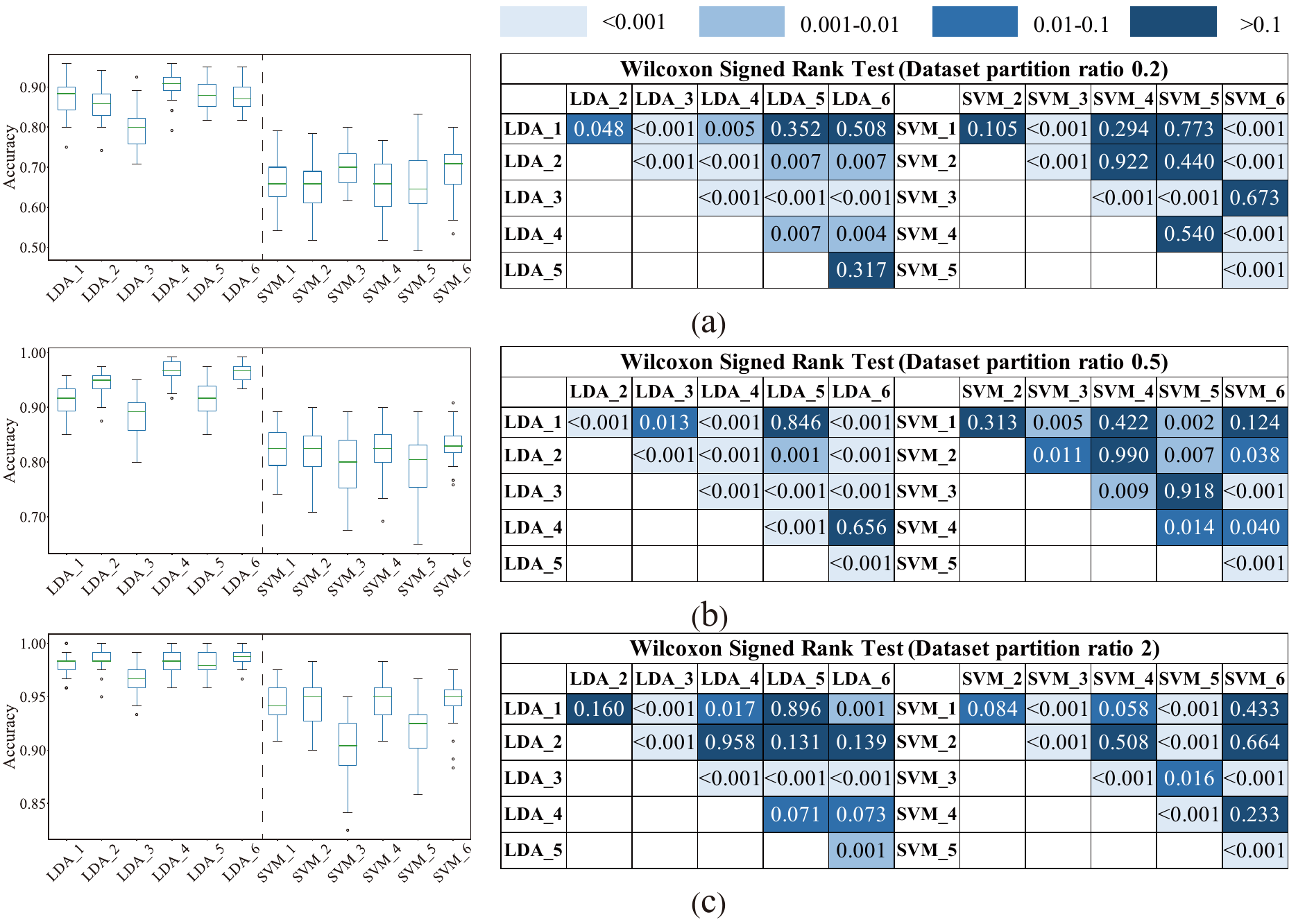}
\caption{\textbf{The classification accuracy and the statistical significance given by Wilcoxon signed-rank tests of six processes in scenario 1 (dataset 1).} The figures denote different ratios of training samples and active samples in scenario 1: 0.2 for 60/300 (a), 0.5 for 120/240 (b), 2 for 240/120 (c).}
\label{Fig:dataset1_scenario1}
\end{figure}

\begin{figure}[htbp]
\centering
\includegraphics[width=\linewidth]{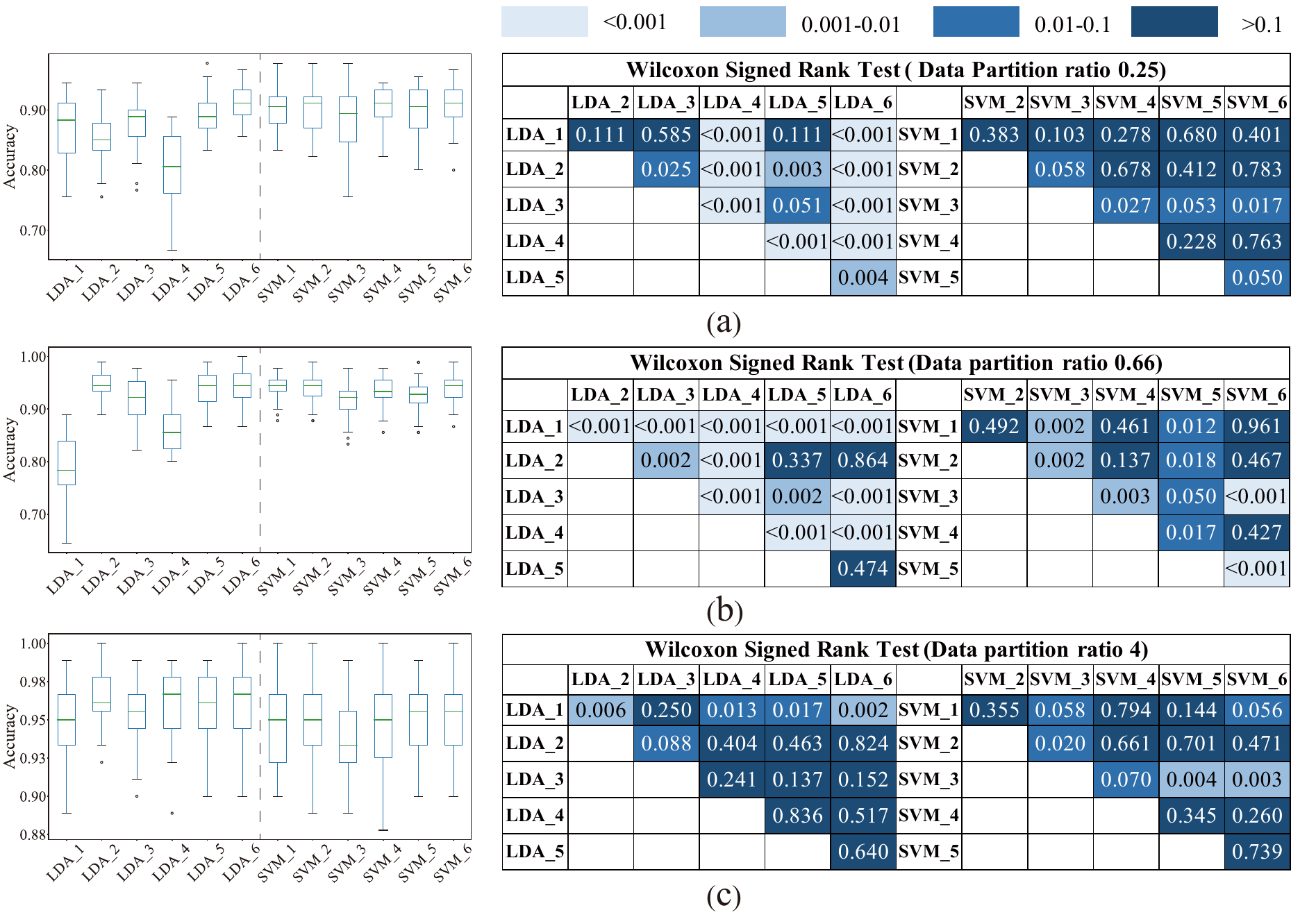}
\caption{\textbf{The classification accuracy and the statistical significance given by Wilcoxon signed-rank tests of six processes in scenario 1 (dataset 2).} The figures denote different ratios of training samples and active samples in scenario 1: 0.25 for 60/240 (a), 0.66 for 120/180 (b), 4 for 240/600 (c).}
\label{Fig:dataset2_scenario1}
\end{figure}

\begin{figure}[htbp]
\centering
\includegraphics[width=\linewidth]{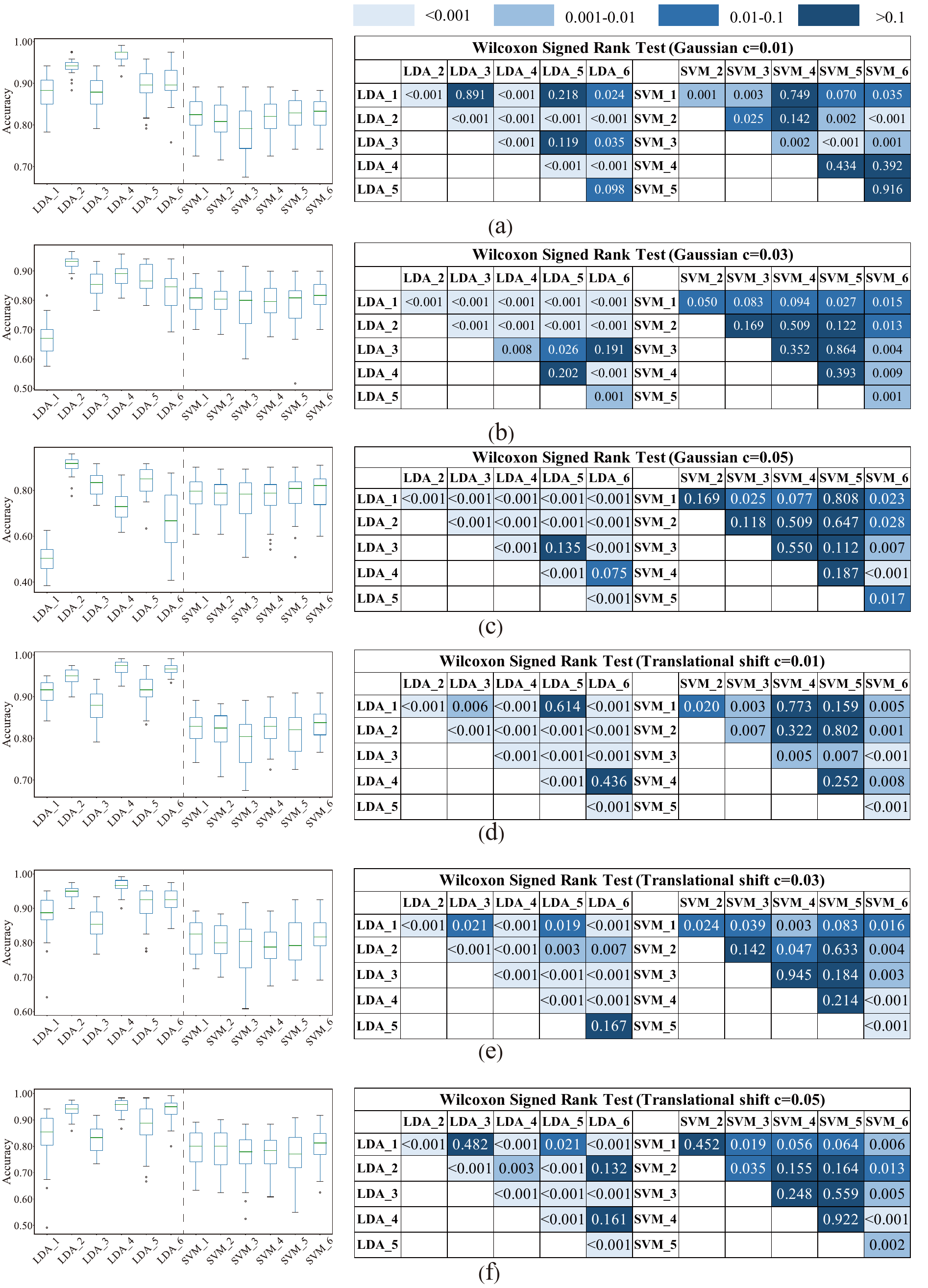}
\caption{\textbf{The classification accuracy and the statistical significance given by Wilcoxon signed-rank tests of six processes in scenario 2 (dataset 1).}The figures denote the three levels of Gaussian noises: c=0.01(a), 0.03(b), 0.05(c) and three levels of translational shift noises: c=0.01(d), 0.03(e), 0.05(f).}
\label{Fig:dataset1_scenario2}
\end{figure}

\begin{figure}[htbp]
\centering
\includegraphics[scale=0.7]{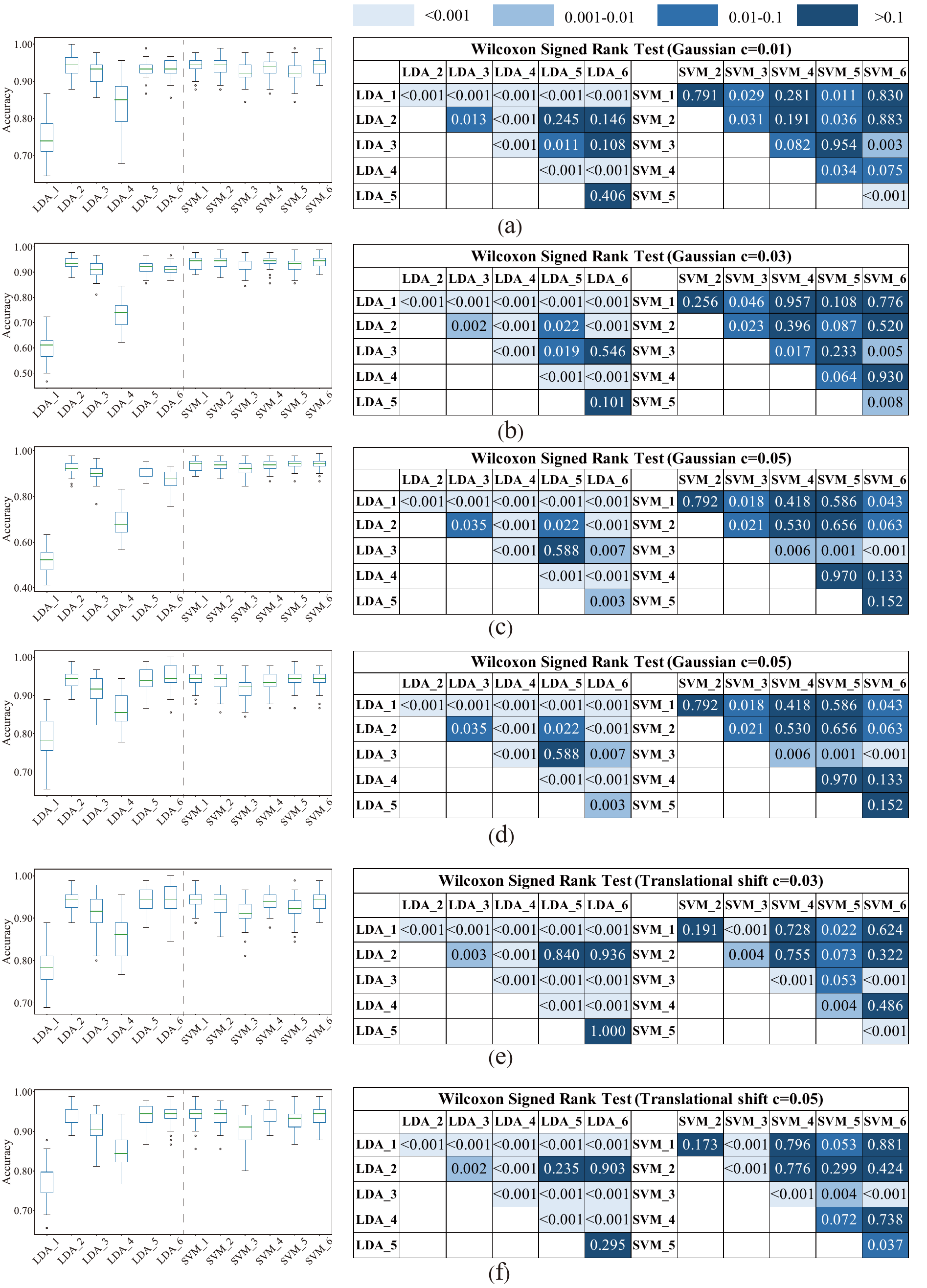}
\caption{\textbf{The classification accuracy and the statistical significance given by Wilcoxon signed-rank tests of six processes in scenario 2 (dataset 2).}The figures denote the three levels of Gaussian noises: c=0.01(a), 0.03(b), 0.05(c) and three levels of translational shift noises: c=0.01(d), 0.03(e), 0.05(f).}
\label{Fig:dataset2_scenario2}
\end{figure}

\newpage
\clearpage
\section*{Acknowledgments}

	The work is supported by the Natural Science Foundation of China (Grant No. 61773342) and the Science Fund for Creative Research Groups of NSFC (Grant No.61621002). In this study, the data was collected at Zhejiang University and the computational modeling was based on a course project for Stanford EE269. The authors appreciate the research equipment and assistance from the State Key Laboratory of Industrial Control Technology, Institute of Cyber-Systems and Control, Zhejiang University.

\section*{Author Contribution}\label{sec:author contribution}
	Li Liu analyzed the data, conducted the experiments and wrote the paper. Xianghao Zhan designed the study pipeline, data augmentation strategies and wrote the paper. Xianghao Zhan, Xiaoqing Guan, Rumeng Wu and Zhan Wang collected the data. Wei Zhang and Mert Pilanci revised the paper. Guang Li, Zhiyuan Luo and You Wang supervised the study. You Wang provided the funding of this study.

\section*{Conflict of interests}
The authors declare no conflict of interests.

\nolinenumbers


\newpage
 \clearpage
\bibliography{ref}
\bibliographystyle{IEEEtran}


\end{document}